\documentclass[11pt]{article}
\usepackage[margin=0.95in]{geometry}
\usepackage{epsfig,ulem,amssymb,soul}

\usepackage{hyperref}
\usepackage{graphicx} 
\usepackage{epstopdf}
\usepackage{amsfonts,amsmath}
\usepackage{bm}
\usepackage{setspace}
\usepackage{comment}
\usepackage{siunitx}
\usepackage{mathrsfs}
\usepackage{multirow}
\usepackage{subcaption}
\usepackage[hang,flushmargin,bottom]{footmisc}
\usepackage{mathtools}
\usepackage{lipsum}
\usepackage{array}
\usepackage{enumerate}
\usepackage{ifthen}
\usepackage[affil-it]{authblk}
\usepackage{cite}
\usepackage{leftidx}
\usepackage{relsize}
\usepackage{braket}
\usepackage{float}

\usepackage[autostyle]{csquotes}
\usepackage[dvipsnames]{xcolor}
\newcommand{\etal}{\textit{et al}.~}

\providecommand{\keywords}[1]{\textbf{\textit{Keywords---}} #1}

\title{Physics and geometry informed neural operator network with application to acoustic scattering}
\date{}

\author[1]{Siddharth Nair}
\author[2]{Timothy F. Walsh}
\author[2]{Greg Pickrell}
\author{Fabio Semperlotti\thanks{To whom correspondence should be addressed. Email: fsemperl@purdue.edu }}
\affil[1]{Ray W. Herrick Laboratories, School of Mechanical Engineering,  Purdue University, West Lafayette, IN 47907, USA}
\affil[2]{Sandia National Laboratories, Albuquerque, NM 87185, USA}

\usepackage{lineno}

\begin{document}
\maketitle

\begin{abstract}
In this paper, we introduce a physics and geometry informed neural operator network with application to the forward simulation of acoustic scattering. The development of geometry informed deep learning models capable of learning a solution operator for different computational domains is a problem of general importance for a variety of engineering applications. To this end, we propose a physics-informed deep operator network (DeepONet) capable of predicting the scattered pressure field for arbitrarily shaped scatterers using a geometric parameterization approach based on non-uniform rational B-splines (NURBS). This approach also results in parsimonious representations of non-trivial scatterer geometries. In contrast to existing physics-based approaches that require model re-evaluation when changing the computational domains, our trained model is capable of learning solution operator that can approximate physically-consistent scattered pressure field in just a few seconds for arbitrary rigid scatterer shapes; it follows that the computational time for forward simulations can improve (i.e. be reduced) by orders of magnitude in comparison to the traditional forward solvers. In addition, this approach can evaluate the scattered pressure field without the need for labeled training data. After presenting the theoretical approach, a comprehensive numerical study is also provided to illustrate the remarkable ability of this approach to simulate the acoustic pressure fields resulting from arbitrary combinations of arbitrary scatterer geometries. These results highlight the unique generalization capability of the proposed operator learning approach. 

\noindent\keywords{Physics-informed DeepONet, Geometry parameterization, Acoustic scattering, NURBS}
\end{abstract}

\section{Introduction}
\label{ssec: Introduction}

Machine learning (ML) has found widespread use in scientific and engineering applications. Among the many different methodologies, deep learning has gained particular prominence thanks to its capability to approximate very complex functions. For example, conventional data-driven deep neural networks (DNNs) have excelled in computer vision and natural language processing applications, which has further motivated the scientific community to test their capabilities in many other areas. When applied to engineering applications, DNNs face significant challenges typically due to the limited availability of sufficiently large and comprehensive databases and to the inherent inability of DNNs to ensure physically consistent results \cite{karniadakis2021physics, karpatne2022knowledge}. A solution to this problem was provided in the form of physics-guided DNNs.

To-date, data-driven DNNs under supervised learning are still a primary choice in scenarios where the governing physical laws are unknown. However, there is undoubtedly a rapidly growing trend towards the use of physics-driven DNNs, especially in engineering applications. Scientific machine learning, as this area of ML is often dubbed, aims at integrating the knowledge of the governing physics into conventional DNNs \cite{karpatne2022knowledge}. In scientific machine learning, the integration of physical laws into the learning algorithm amounts to introducing appropriate biases in the DNN that can guide the network prediction towards physically consistent solutions. 
There are multiple approaches to embed the physical laws describing the problem at hand.
The first approach is to introduce this physical bias via additional observational data that embody the underlying physics. A variety of supervised and semi-supervised learning-based networks trained with observational data are particularly valuable in situations where no direct mathematical relationship exists but some prior physical data can be embedded into the DNN. This is especially relevant for many types of inverse problems including, but not limited to, property identification and design optimization with applications to mechanics \cite{white2019multiscale, zhang2021multi, pestourie2020active}, materials \cite{guo2021artificial}, and acoustics \cite{wu2022physics, nair2023grids}. The major limitation of this approach is the cost associated with the generation of a large volume of training data to reinforce the physical bias \cite{karniadakis2021physics}.
The second approach is through tailored DNN architectures that embed the prior physical information to bias the network prediction, such as DNN architectures encoding multiscale features \cite{liu2020multi}, symmetries and energy conservation \cite{mattheakis2019physical}, and high frequency components \cite{cai2020phase}. However, this approach is constrained by its complexity to scale, and therefore, it is typically used to implement simple symmetry groups that are known a priori \cite{karniadakis2021physics}. A third approach is to introduce learning bias by enforcing the governing physics through appropriate loss functions and constraints in the DNN. One of the most extensively studied and effective examples of this is the concept of physics-informed neural network (PINN).  

Raissi $\etal$ \cite{raissi2019physics} introduced the concept of PINNs as an alternative method to find the numerical solution of partial differential equations (PDEs). As the dynamic behavior of systems in most engineering applications is often described using conservation and constitutive laws expressed as a system of PDEs, PINNs find application in a wide range of domains including, as an example, solid mechanics \cite{shukla2020physics, samaniego2020energy}, fluid mechanics \cite{raissi2020hidden, jin2021nsfnets}, heat transfer \cite{zobeiry2021physics}, biomedical engineering \cite{yazdani2020systems}, inverse design \cite{lu2021physics}, and photonics \cite{chen2020physics}. The major advantage of using PINNs lies in the ability to train without labeled datasets and in learning to predict physically consistent results from governing equations. In addition, unlike the conventional simulation approach of finite element (FE) analysis, the PINNs have lower discretization dependence. A fundamental extension to this class of problems consists in solving a system of PDEs for a range of conditions (as opposed to just a single one) like different input variables, boundary conditions, and domain geometries. In order to solve these so-called parametric PDE problems \cite{wang2021learning}, the associated ML model must be able to learn the solution operator that maps a range of inputs to the corresponding solution of the underlying PDE system. However, addressing this problem using finite elements (FE) methods and PINNs proves ineffective. Both methods incur significant computational cost due to the need to perform either independent simulations (in FE methods) or trainings (in PINNs) for each variation of the input parameters.

Recently, Lu \etal \cite{lu2021learning} introduced a novel DNN architecture for operator learning based on the universal approximation theorem for operators; this approach is often referred to as deep operator network (DeepONet). DeepONets are capable of approximating nonlinear operators by mapping parametric input functions to solution functions. This operator learning network has been extensively studied to develop wide range of applications involving boundary layer instabilities in fluid mechanics \cite{di2021deeponet}, organ simulations in biomedical engineering \cite{yin2022simulating}, full waveform inversion in wave propagation \cite{zhu2023fourier}, and uncertainty quantification in the energy sector \cite{moya2023deeponet}, to name a few.
In their conventional form, a major drawback of DeepONets is the need for a comprehensive training dataset encompassing the complete distribution of the underlying physical behavior across parameters to capture the function-function mappings. To address this limitation, Wang \etal \cite{wang2021learning} introduced a new deep operator network concept called the physics-informed DeepONet. In a broad sense, physics-informed DeepONet or PI-DeepONet is an operator learning network that integrates the abilities of PINN within DeepONets without inheriting the corresponding limitations. More specifically, PI-DeepONet is capable of solving a system of parametric PDEs by learning solution operators that can map between functional spaces instead of vector spaces, as it is the case for PINNs. Moreover, unlike PINNs that require undergoing new training for each parameter, once trained, the PI-DeepONet can approximate solutions for any independent parameter value, within a continuous parameter range, at orders of magnitude faster than the traditional PDE solvers. 

Early work in this field has introduced PI-DeepONets to approximate mappings between parameterized coefficients in PDEs and their solution operators \cite{lu2021learning, koric2023data}. Subsequent research has also developed PI-DeepONets incorporating parametric boundary and initial conditions \cite{goswami2022physics, wang2023long}. 
However, the existing literature includes limited studies that develop PI-DeepONet capable of accurately approximating the mapping between a geometry function space (i.e. a space capable of capturing a range of arbitrary shapes, which is referred to as a variable computational domain) and the corresponding solution operator. It is important to note that unlike parametric coefficients and boundary conditions, the concept of a variable computational domain is not a direct parametric PDE problem, as there is 
no explicit PDE parameter that accounts for the variation in the domain shape. Consequently, this makes the approximation of the solution operator even more complex.
Although \cite{lu2021learning} briefly studies a related example by solving the Eikonal equation, their approach provides the 2D Cartesian coordinates of the domain as input therefore limiting the prediction capabilities to one family of basic shapes. In other words, the PI-DeepONet from this study can only accurately predict solution operators for a predefined set of shapes. This limitation primarily arises from the inability to represent and capture geometries of various shapes and sizes using a single set of geometry parameters. 
To address these limitations, it is critical to adopt a uniform geometry representation that accommodates a wide range of shapes without significantly increasing the dimensionality and computational complexity of the problem.

Geometry parameterization via non-rational B-splines (NURBS) offers a parsimonious dimensional representation of complex geometries compared to the conventional Cartesian coordinates representation. Among the various existing geometry parameterization techniques, NURBS can effectively represent any basic or free-form shape. The ability to represent a large variety of shapes via a relatively small and fixed number of parameters makes NURBS an effective tool to represent arbitrary geometries \cite{ma1998nurbs, saini2017nurbs}. Traditionally, NURBS has been an integral part of FE based simulation tools \cite{hughes2005isogeometric}. In recent years, the integration of DNNs with NURBS-based isogeometric parameterization has been used to solve forward problems in biomechanics \cite{balu2019deep,zhang2021simulating}. More recently, Nair \etal \cite{nair2023grids} and Wu \etal \cite{wu2022physics} developed DNN-based hybrid learning models for inverse material design with NURBS geometry parameterization. However, it appears that this concept has not been integrated into physics-informed DNNs (either PINNs or PI-DeepONet) to achieve accurate yet parsimonious geometry representation.

This study presents a physics-informed deep operator network model for variable computational domains, hence addressing some important limitations of existing operator networks. The effectiveness of this operator learning model is demonstrated by application to acoustic scattering problems. More specifically, a NURBS-based geometry parameterization integrated PI-DeepONet is developed and is capable of learning a solution operator that can accurately approximate the scattered pressure field generated by an arbitrary-shaped rigid body scatterer almost in real time. In the remainder of this paper, we will refer to this geometry-aware operator learning model as \textbf{P}hysics- and \textbf{G}eometry-\textbf{I}nformed \textbf{Deep} \textbf{O}perator \textbf{Net}work or \textbf{PGI-DeepONet}. The contributions of this study are two-fold: 
\begin{enumerate}
    \item It provides the formulation of the PGI-DeepONet that leverages NURBS-based geometry parameterization to represent variable computational domains. Unlike the existing PI-DeepONets, the proposed PGI-DeepONet is capable of approximating the solutions for computational domains embedded with NURBS-based geometry of arbitrary shape and size.
    \item The development of a deep operator network to simulate acoustic scattering problems without relying on any labeled training dataset, but instead by enforcing the underlying physics through the governing PDEs. The PGI-DeepONet also offers improved generalization performance. In contrast to the existing methods, like FE and PINN, which require re-evaluation for each independent scatterer shape, the trained PGI-DeepONet can speed up the simulation of scattering problems involving arbitrary-shaped scatterers by orders of magnitude. The reduction of the computational time employed by the forward solvers is of significance as there are many practical applications that require multiple evaluations of the forward problem, such as inverse design based on iterative solvers.
\end{enumerate}

The paper is organized as follows. In \S\ref{ssec: theoretical_preliminaries}, we introduce the fundamental concepts and properties of PINN and DeepONet. Further, \S\ref{ssec: problem_statement} presents the general problem setup and introduces PGI-DeepONet model for acoustic scattering. In \S\ref{ssec: Methodology}, we elaborate on the implementation of the PGI-DeepONet, with specific attention to the details of the network architecture, geometry preparation, and training procedure. Finally, \S\ref{ssec: Results} reports the results of the trained PGI-DeepONet to the acoustic scattering problem.

\section{Some preliminary concepts of deep learning }
\label{ssec: theoretical_preliminaries}

This section provides a review of some key elements involved in the development of the proposed physics-informed operator learning network. More specifically, we discuss two fundamental deep learning concepts: 1) physics-informed neural networks (PINNs), and 2) deep operator networks (DeepONets). These concepts are crucial to integrate the physics of the problem (via governing equations) into deep neural networks and to understand the ability of deep neural networks to learn nonlinear operator mappings.

\subsection{Physics-informed neural network (PINN)}
\label{ssec: PINN_preliminaries}

Physics-informed neural networks (PINNs) belong to a class of deep learning algorithms that can integrate prior physical information about the problem with the intent of improving the performance of the learning algorithms. PINNs force the deep learning algorithm to learn the mathematical framework of the problem in the form of governing partial differential equations (PDEs) and constitutive equations.

In order to understand the implementation of PINNs, consider a PDE defined on a 2D domain $\Omega_0 \subset \mathbb{R}^2$ in the following form 
\begin{equation}
    \label{eqn: PINN_PDE}
    \mathcal{N} (\phi, \frac{\partial \phi}{\partial x}, \frac{\partial \phi}{\partial y}, \frac{\partial^2 \phi}{\partial x \partial x},...\frac{\partial^2 \phi}{\partial y \partial y};\textbf{x}) = f(\textbf{x}) ~~~~~~~ \textbf{x}=(x,y) \in \Omega_0
\end{equation}
with boundary conditions on $\partial \Omega_0$
\begin{equation}
    \label{eqn: PINN_BC}
     \mathcal{B} (\phi, \frac{\partial \phi}{\partial \textbf{n}}; \textbf{x})= g(\textbf{x}) ~~~~~~~ \textbf{x}=(x,y) \in \partial \Omega_0
\end{equation}
where $\phi$ is the solution, $\textbf{n}$ is the surface normal, $f$ is the source function, $g$ is the boundary function, $\mathcal{N}$ represents the governing PDE, and $\mathcal{B}$ represents boundary conditions. Depending on the application, $\mathcal{B}$ can represent either Dirichlet, Neumann, or mixed (Robin) boundary conditions.

In a PINN framework, the unknown solution $\phi$ is computationally predicted by a neural network via a parameter set $\theta=[\textbf{W}, \textbf{b}]$, where \textbf{W} and \textbf{b} are the set of weights and biases of the neural network, respectively. The PINN approximation of the solution is $\hat{\phi}$ such that $\hat{\phi} \approx \phi$.
Moreover, the PINN learns to predict $\hat{\phi}$ by finding the optimal $\theta$ that minimizes a loss function $\mathcal{L}(\theta)$
\begin{equation}
     \label{eqn: PINN_theta}
     \theta^* = \arg \min_{\theta} \mathcal{L}(\theta) 
\end{equation}
where $\mathcal{L}(\theta)$ is a combination of $\mathcal{L}_\mathcal{B}(\theta)$ and $\mathcal{L}_\mathcal{N}(\theta)$ with their respective weighting factors $w_{\mathcal{B}}$ and $w_{\mathcal{N}}$
\begin{equation}
     \label{eqn: PINN_loss}
     \mathcal{L}(\theta) = w_{\mathcal{B}}\mathcal{L}_{\mathcal{B}}(\theta) + w_{\mathcal{N}}\mathcal{L}_{\mathcal{N}}(\theta)
\end{equation}
and
\begin{equation}
\begin{split}
     \label{eqn: PINN_LossPDEBC}
     \mathcal{L}_{\mathcal{B}}(\theta) &= \frac{1}{N_\mathcal{B}} \sum^{N_\mathcal{B}}_{i=1} \Big|\mathcal{B} (\hat{\phi}, \frac{\partial \hat{\phi}}{\partial \textbf{n}}; \textbf{x}) - g(\textbf{x}) \Big|^2 \\
     \mathcal{L}_{\mathcal{N}}(\theta) &=  \frac{1}{N_\mathcal{N}} \sum^{N_\mathcal{N}}_{i=1} \Big|\mathcal{N} (\hat{\phi}, \frac{\partial \hat{\phi}}{\partial x}, \frac{\partial \hat{\phi}}{\partial y}, \frac{\partial^2 \hat{\phi}}{\partial x \partial x},...\frac{\partial^2 \hat{\phi}}{\partial y \partial y};\textbf{x}) - f(\textbf{x}) \Big|^2
\end{split}
\end{equation}
where $\mathcal{L}_\mathcal{B}(\theta)$ and $\mathcal{L}_\mathcal{N}(\theta)$ are the mean square errors (MSE) of the residual of the boundary condition (Eq.~\ref{eqn: PINN_BC}) and of the governing PDE (Eq.~\ref{eqn: PINN_PDE}), respectively. In addition, $N_\mathcal{B}$ is the number of training points on $\partial \Omega_0$ and $N_\mathcal{N}$ is the number of collocation points in $\Omega_0$.

In summary, PINNs can learn to predict the physical solution by approximating solutions satisfying the mathematical equations governing the problem. The two major advantages of using PINNs in contrast to conventional DNNs are: 1) PINNs can be trained without the use of labeled datasets, and 2) PINNs can learn to predict physically consistent results. While these points highlight the potential of PINNs for both forward and inverse simulations, it is important to note that they essentially operate as domain-specific PDE solvers. In other terms, PINNs are trained on equations that correspond to specific physical domains and sets of boundary conditions. Consequently, they need to undergo a new training phase every time the geometry of the physical domains, boundary conditions, or parameters of the governing PDE are changed. This limitation implies that while PINNs are capable of function approximations, they are incapable of learning domain generalizable operator solutions.

\subsection{Deep operator network (DeepONet)}
\label{ssec: DeepONet_preliminaries}

\begin{figure}[h!]
	\centering
	\includegraphics[width=1.0\linewidth]{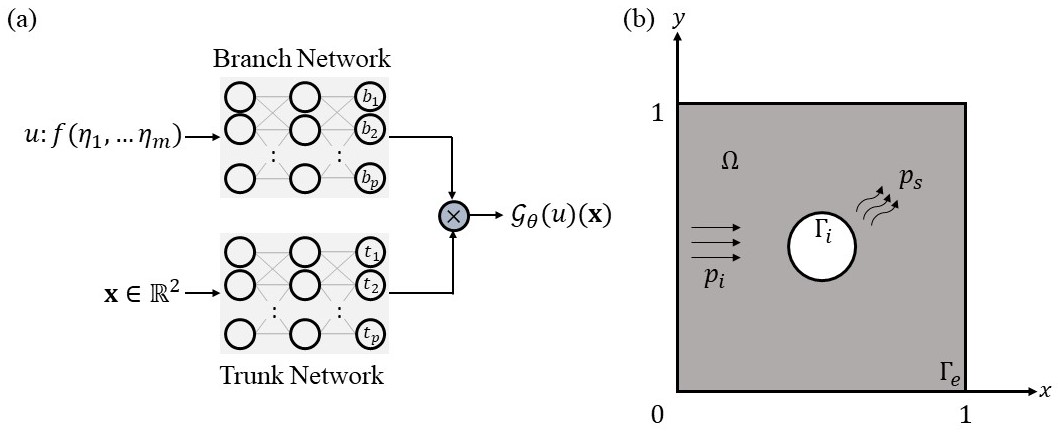}
	\caption{Schematic illustrates a sample (a) DeepONet architecture with branch and trunk networks. Here, $u$ represents an input parameterization function evaluated at $m$ discrete points and $\textbf{x}$ describes the spatial coordinates in a 2D domain, while $\mathcal{G}_\theta$ represents the learned operator that maps the inputs $u$ and $\textbf{x}$ to the corresponding output $\mathcal{G}_{\theta}(u)(\textbf{x})$. (b) 2D square acoustic domain $\Omega$ of size $[0,1]~m \times [0,1]~m$ with internal boundary $\Gamma_i$ and external boundary $\Gamma_e$. This also highlights the incident pressure ($p_i$) and scattered pressure ($p_s$) inside the domain.}
	\label{fig: AcousticF_DeepONet}
\end{figure}

Neural networks are commonly identified as universal function approximators, but an equally significant feature is the property of neural networks to act as universal operator approximators \cite{chen1995universal}. Recently, Lu \etal \cite{lu2021learning} proposed a class of data-driven deep operator networks called DeepONets for accurate operator approximation. The common DeepONets architecture consists of two sub neural networks as shown in Fig.~\ref{fig: AcousticF_DeepONet}(a): 1) a branch network to encode the information related to the input parameterization function, and 2) a trunk network to encode the information related to the discrete spatial coordinates to evaluate the output. More specifically, the DeepONet architecture can learn an operator $\mathcal{G}_{\theta}$ that can map the spatial domain coordinates $\textbf{x}$ (input to the trunk network) to the corresponding output fields (such as temperature, pressure, velocity, etc.) across a range of parameters described by the input parameterization function $u$ (input to the branch network). These input parameters may include, but are not limited to, variable coefficients, source terms, boundary conditions, or the shape of the domain. In order to make use of the input $u$, the function is evaluated at $m$ discrete values $\{\eta_1,\eta_1,...\eta_m\}$ in the finite-dimensional space as $[u(\eta_1), u(\eta_2), ... u(\eta_m)]$. Based on this, the solution operator $\mathcal{G}$ can be expressed as 
\begin{equation}
\begin{split}
     \label{eqn: G0_operator}
     \mathcal{G}_{\theta}(u)(\textbf{x}) &= \sum^p_{i=1} b_i t_i \\
     &= \sum^p_{i=1} b_i(u(\eta_1), u(\eta_2), ... u(\eta_m))t_i(\textbf{x})
\end{split}
\end{equation}
where $b_i$ and $t_i$ for $i=1,2,...p$ are the outputs of the branch and trunk networks, respectively. The interested reader can refer to \cite{lu2021learning} for extensive details on the derivation of Eq.~\ref{eqn: G0_operator}. The data-driven DeepONet can now be trained to satisfy the following loss function
\begin{equation}
     \label{eqn: loss_DeepONet}
     \mathcal{L}(\theta) = \frac{1}{N} \sum^N_{i=1} |s_i(\textbf{x}) - \mathcal{G_{\theta}}(u_i)(\textbf{x})|^2
\end{equation}
where $s$ represents true labeled data for $N$ number of training samples. More specifically, the DeepONet learns to predict $\mathcal{G}_{\theta}(u)(\textbf{x})$ by finding the optimal network parameters $\theta$ that minimize $\mathcal{L}(\theta)$ in Eq.~\ref{eqn: loss_DeepONet}.

Although DeepONets are capable of learning operators, they still require large labeled datasets to train the network for generalized solutions. This trend is in contrast with most of the fundamental engineering applications governed by well-established mathematical principles that often suffer from a scarcity of available training data. This data scarcity can be counteracted by developing operator learning approaches that are primarily physics-driven, hence leading to the introduction of the field of physics-informed (PI) DeepONets \cite{wang2021learning}. In a general sense, the PI-DeepONet can be seen as a network architecture that integrates the key attributes of PINNs and DeepONets. The PI-DeepONet is capable of learning the solution operator that maps the complete range of parameter input to the corresponding solutions of the appropriate parameterized system of equations. 

While this section motivated and introduced the general concept of PI-DeepONet, the following section will elaborate on the specific details regarding the development and implementation of PI-DeepONet tailored to address the forward problem of acoustic scattering via arbitrarily shaped rigid body scatterers.

\section{Problem definition}
\label{ssec: problem_statement}

To illustrate the PI-DeepONet concept, we choose a benchmark problem based on rigid body acoustics. In the following, the general problem setup is first introduced. Then, a brief overview of the well-established concept of non-uniform rational B-Spline (NURBS) is provided. Finally, the PI-DeepONet model for rigid body acoustic scattering is presented.

\subsection{Rigid body acoustic scattering}
\label{ssec: Acoustic_scattering}

Consider the classical problem of a harmonic plane wave impinging on a sound hard scatterer defined in a 2D domain $\Omega \subset \mathbb{R}^2$  with $\mathbf{x}=(x,y)$, as shown in Fig.~\ref{fig: AcousticF_DeepONet}(b). Assume without loss of generality that the incoming wave $p_i = p_0e^{-i\textbf{k} \cdot \textbf{x}}$ has a unit amplitude ($p_0=1~Pa$). The problem is governed by the Helmholtz equation
\begin{equation}
    \label{eqn: Helmholtz_equation}
    \mathcal{N} (p_s,\textbf{x}):=\nabla^2 p_s(\textbf{x}) + k^2 p_s(\textbf{x}) = 0, ~~~~~~~ \textbf{x} \in \Omega
\end{equation}
The sound hard scatterer is represented by a rigid internal boundary described by the Neumann boundary condition on $\Gamma_{i}$ 
\begin{equation}
    \label{eqn: rigid_BC}
     \mathcal{B}_i (p_s,\textbf{x}):=\frac{\partial p_s(\textbf{x})}{\partial \textbf{n}}-ike^{-i\textbf{k} \cdot \textbf{x}}=0, ~~~~~~~ \textbf{x} \in \Gamma_{i}
\end{equation}
with an impedance condition on the external boundary $\Gamma_e$
\begin{equation}
    \label{eqn: ext_BC}
     \mathcal{B}_e (p_s,\textbf{x}):=\frac{\partial p_s(\textbf{x})}{\partial \textbf{n}} + ikp_s(\textbf{x})=0, ~~~~~~~ \textbf{x} \in \Gamma_{e}
\end{equation}
where $p_s$ is the scattered pressure, the wavenumber $\textbf{k} = \frac{2\pi f}{c_s} \hat{\textbf{e}}_k$ is a function of the frequency $f$, $c_s$ is the speed of sound, $\hat{\textbf{e}}_k=[1,0]$ is the unit normal vector, and $\textbf{n}$ is the surface normal. The differential operator $\mathcal{N}$ represents the governing PDE, while the boundary operators $\mathcal{B}_i$ and $\mathcal{B}_e$ represent the internal and external boundary conditions, respectively. The Eqs.~(\ref{eqn: Helmholtz_equation})-(\ref{eqn: ext_BC}) completely describe the rigid body acoustic scattering problem in an infinite domain. In addition, $p_s= \mathfrak{Re}(p_s) + i\mathfrak{Im}(p_s)$ is the general solution of Eq.~(\ref{eqn: Helmholtz_equation}) for what concerns the pressure scattered by the rigid scatterer. It follows that, given a certain incident wave pressure $p_i$, the solution to Eq.~(\ref{eqn: Helmholtz_equation}) is controlled by the shape of the rigid scatterer boundary $\Gamma_i$. 

When using physics-based simulation approaches for acoustic scattering problems, such as FE method or even deep learning-based PINNs, the solution $p_s$ minimizes a domain-specific functional based on the governing equation and the boundary conditions. If the parameters describing the problem change (e.g. different boundary conditions, input loads, etc.), the finite element model must be re-evaluated and PINN must be re-trained. The PI-DeepONet proposed in this study will allow simulating the problem for any arbitrary shape without the need for any additional training. The result will be an extremely computationally efficient approach to forward simulations. 

At the same time, a significant issue with arbitrary geometry representation is that the same shape can be described by different sets of parameters. As most DNN architectures have difficulty training with high dimensional data, it is critical to use representations that can capture shapes accurately while using the minimum number of geometry parameters. Therefore, the choice of the geometry parameterization approach becomes central to achieve accurate and efficient solutions via neural networks.

\subsection{Shape representation using NURBS}
\label{ssec: NURBS_importance}

This section explores the possibility to use non-uniform rational B-spline (NURBS) to convert the high-dimensional parameter space of shapes into an accurate yet low-dimensional representation. Note that, in this context, dimensionality refers to the number of parameters required to describe a scatterer shape.

NURBS has found extensive use in the parametric representation of various 2D geometries because it allows the mathematical description of multiple complex shapes via a fixed set of parameters (i.e. control points ($\textbf{C}$) and corresponding weights ($w$)). Of course, given that NURBS provides an approximation of the true shape, it will naturally introduce some approximation errors. However, previous studies \cite{nair2023grids} suggest that the approximation error can be limited by a proper choice of the NURBS parameters.

In NURBS parametric representation of shapes, we define a curve of degree $\tilde{k}$ with $n+1$ control points $[\textbf{C}_0, \textbf{C}_1,...\textbf{C}_n]$ associated with $n+1$ weights $[w_0, w_1,...w_n]$, and a knot vector $t=[t_0, t_1,...t_{\tilde{m}}]$ with $\tilde{m}+1$ knots, where $\tilde{m}=n+\tilde{k}+1$ \cite{piegl1996nurbs, hughes2005isogeometric}. The knot vector is a non-decreasing sequence of geometric coordinates ($t_i$) that splits the B-splines into non-uniform piecewise functions. The NURBS curve $\tilde{\textbf{C}}$, parameterized using $u$, is defined as follows \cite{piegl1996nurbs}
\begin{equation}
    \label{eqn: NURBS_CP}
    \tilde{\textbf{C}}(u) = \frac{\sum_{i=0}^n w_i N_{i,\tilde{k}}(u)\textbf{C}_i}{\sum_{j=0}^n w_j N_{j,\tilde{k}}(u)}
\end{equation}
where $\textbf{C}_i=[C^x_i, C^y_i]$, $C^x_i$ and $C^y_i$ are the $x$ and $y$ coordinates of the $i^{th}$ control point, and $N_{i,\tilde{k}}(u)$ is the associated B-spline basis function.
At this point, $N_{i,\tilde{k}}(u)$ is evaluated recursively using Cox-de Boor recursion formula as follows
\begin{equation}
    \label{eqn: NURBS_basis}
    N_{i,\tilde{k}}(u) = \frac{(u-t_i)N_{i,\tilde{k}-1}(u)}{t_{i+\tilde{k}}-t_i} + \frac{(t_{i+\tilde{k}+1}-u)N_{i+1,\tilde{k}-1}(u)}{t_{i+\tilde{k}+1}-t_{i+1}}
\end{equation}
and 
\begin{equation}
    \label{eqn: NURBS_basis0}
    N_{i,0}(u)=
    \begin{cases}
        1 & t_i \leq u \leq t_{i+1}\\
        0 & \text{otherwise}
    \end{cases}
\end{equation}
Eq.~(\ref{eqn: NURBS_CP}) defines NURBS curve $\tilde{\textbf{C}}$ as a function of NURBS parameters $\textbf{C}$ and $w$. The interested reader can refer to \cite{nair2023grids} for further details on the NURBS-based geometry parameterization approach. 
The use of the NURBS approach (described by Eqs.~(\ref{eqn: NURBS_CP})-(\ref{eqn: NURBS_basis0})) offers notable advantages for the acoustic problem at hand: 1) NURBS can represent complex 2D scatterer shapes within a highly reduced parametric space, 2) the ability to describe arbitrary 2D shapes with a concise set of parameters can significantly reduce the computational cost associated with the DNN training, and 3) NURBS enables the development of a generalized DNN architecture with a fixed number of input parameters for shape representation. 


\subsection{Physics-informed operator learning for acoustic scattering}
\label{ssec: PGI_DeepONet}

In this section, we discuss the development of a physics-informed deep operator learning framework to simulate the forward acoustic scattering problem presented in \S\ref{ssec: Acoustic_scattering}. The framework also integrates a NURBS-based geometry representation (\S\ref{ssec: NURBS_importance}) to generalize the network performance to arbitrary 2D shapes. 

\begin{figure}[h!]
	\centering
	\includegraphics[width=1.0\linewidth]{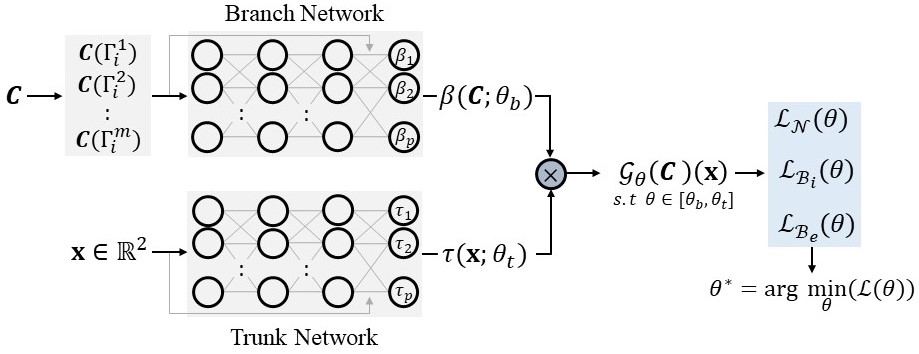}
	\caption{Schematics of the neural network architecture of the PGI-DeepONet. $\textbf{C}$ represents an input NURBS parameterization function evaluated for $m$ discrete shapes $\Gamma_i$, and $\textbf{x}$ describes the spatial coordinates forming the corresponding acoustic domain. In addition, $\beta(\textbf{C}; \theta_b)$ are the coefficients of the branch network, $\tau(\textbf{C}; \theta_t)$ are the coefficients of the trunk network, and $\mathcal{G}_\theta$ is the learned operator that maps the inputs $\textbf{C}$ and $\textbf{x}$ to the corresponding output $\mathcal{G}_{\theta}(u)(\textbf{x})$.}
	\label{fig: Acoustic_PIDeepONet}
\end{figure}

Building on the fundamental concept of PI-DeepONet introduced in \S\ref{ssec: DeepONet_preliminaries}, we present an operator network called PGI-DeepONet that combines physics-informed operator learning with NURBS-based geometry parameterization. Similar to conventional DeepONets, the PGI-DeepONet architecture consists of two sub-neural networks, denominated the branch and trunk networks (Fig.~\ref{fig: Acoustic_PIDeepONet}). The parameterized low-dimensional shape representation of the rigid scatterer boundary $\Gamma_i$, achieved through the NURBS control points $\textbf{C}$, serves as the input function to the branch network. More specifically, in order to learn to predict over a continuous range of parameterized geometries, a set of input $\textbf{C}$ (which represent a discrete collection of NURBS shapes $\{\Gamma^k_i\}^m_{k=1}$) must be provided to the branch network. On the other hand, the spatial coordinates $\textbf{x}=(x,y)$ within the acoustic domain serve as the input to the trunk network. Subsequently, based on the learned basis coefficients $\beta$ and $\tau$ from the branch and trunk networks, respectively, the PGI-DeepONet output $\mathcal{G}_{\theta}$ is evaluated as follows
\begin{equation}
\begin{split}
     \label{eqn: G_operator}
     \mathcal{G}_{\theta}(\textbf{C})(\textbf{x}) &= \sum^p_{j=1} \beta_j (\textbf{C}; \theta_{\textbf{b}}) \tau_j (\textbf{x}; \theta_{\textbf{t}}) \\
     &= \sum^p_{j=1} \beta_j(\textbf{C}(\Gamma^1_i), \textbf{C}(\Gamma^2_i), ... \textbf{C}(\Gamma^m_i); \theta_{\textbf{b}}) \tau_j(\textbf{x}; \theta_{\textbf{t}})
\end{split}
\end{equation}
where, the coefficients $\beta_j (\textbf{C}; \theta_{\textbf{b}})$ are functions of the NURBS control points (\textbf{C}), while $\tau(\textbf{x}; \theta_{\textbf{t}})$ is the function of the spatial coordinates ($\textbf{x}$). In addition, $\theta=[\theta_b, \theta_t]$ denotes the collection of all trainable parameters of the PGI-DeepONet, where $\theta_b$ and $\theta_t$ represent the set of trainable parameters of the branch and trunk networks, respectively.

The physics governing the problem is enforced via the loss function of the network. More specifically, the PGI-DeepONet solution $\mathcal{G}_{\theta}$ satisfies the governing physics of the rigid body acoustic scattering problem by directly enforcing Eqs.~(\ref{eqn: Helmholtz_equation})-(\ref{eqn: ext_BC}) in the loss function. Moreover, the PGI-DeepONet learns to predict accurate $\mathcal{G}_{\theta}$ by finding the optimal $\theta$ that minimizes the physics-driven loss function $\mathcal{L}(\theta)$ as follows

\begin{equation}
     \label{eqn: loss_PIDeepONet}
     \mathcal{L}(\theta) = \underbrace{w_{\mathcal{B}_i}\mathcal{L}_{\mathcal{B}_i}(\theta)}_{Internal~BC} + \underbrace{w_{\mathcal{B}_e}\mathcal{L}_{\mathcal{B}_e}(\theta)}_{External~BC} + \underbrace{w_{\mathcal{N}}\mathcal{L}_{\mathcal{N}}(\theta)}_{Governing~PDE}
\end{equation}
where $w_{\mathcal{B}_i} = w_{\mathcal{B}_e} = w_{\mathcal{N}} =1$ are the weighting factors and
\begin{equation}
\begin{split}
     \label{eqn: loss_PIDeepONet_details}
     \mathcal{L}_{\mathcal{B}_i}(\theta) &=\frac{1}{m N_{\Gamma_i}} \sum^{m}_{j=1} \sum_{\textbf{x} \in \Gamma_i}|\mathcal{B}_i( \mathcal{G}_{\theta}(\textbf{C}_j)(\textbf{x}_j);\theta)|^2 \\
     &= \frac{1}{m N_{\Gamma_i}} \sum^{m}_{j=1} \sum_{\textbf{x} \in \Gamma_i} \Big|\frac{\partial  \mathcal{G}_{\theta}(\textbf{C}_j)(\textbf{x}_j)}{\partial \textbf{n}}-ike^{-i\textbf{k} \cdot \textbf{x}_j}  \Big|^2 \\
     \mathcal{L}_{\mathcal{B}_e}(\theta) &= \frac{1}{m N_{\Gamma_e}} \sum^{m}_{j=1} \sum_{\textbf{x} \in \Gamma_e}|\mathcal{B}_e( \mathcal{G}_{\theta}(\textbf{C}_j)(\textbf{x}_j);\theta)|^2 \\
     &= \frac{1}{m N_{\Gamma_e}} \sum^{m}_{j=1} \sum_{\textbf{x} \in \Gamma_e} \Big|\frac{\partial \mathcal{G}_{\theta}(\textbf{C}_j)(\textbf{x}_j)}{\partial \textbf{n}} + ik\mathcal{G}_{\theta}(\textbf{C}_j)(\textbf{x}_j)\Big|^2 \\
     \mathcal{L}_{\mathcal{N}}(\theta) &= \frac{1}{m N_{\Omega}} \sum^{m}_{j=1} \sum_{\textbf{x} \in \Omega}|\mathcal{N}(\mathcal{G}_{\theta}(\textbf{C}_j)(\textbf{x}_j);\theta)|^2 \\
     &= \frac{1}{m N_{\Omega}} \sum^{m}_{j=1} \sum_{\textbf{x} \in \Omega} \Big| \nabla^2 \mathcal{G}_{\theta}(\textbf{C}_j)(\textbf{x}_j) + k^2 \mathcal{G}_{\theta}(\textbf{C}_j)(\textbf{x}_j)  \Big|^2
\end{split}
\end{equation}
where $\mathcal{L}_{\mathcal{B}_i}$, $\mathcal{L}_{\mathcal{B}_e}$ and $\mathcal{L}_{\mathcal{N}}$ are the mean square errors (MSE) of the residual of the internal boundary condition (Eq.~\ref{eqn: rigid_BC}), the external boundary condition (Eq.~\ref{eqn: ext_BC}), and the governing PDE (Eq.~\ref{eqn: Helmholtz_equation}), respectively. In addition, $N_{\Gamma_i}$ and $N_{\Gamma_e}$ denote the number of boundary points on $\Gamma_i$ and $\Gamma_e$, while $N_\Omega$ represents the number of collocation points in $\Omega$. In addition, a comparison between the Eqs.~(\ref{eqn: Helmholtz_equation})-(\ref{eqn: ext_BC}) and residual terms in Eq.~\ref{eqn: loss_PIDeepONet} highlights that the PGI-DeepONet approximation of the solution operator for given domain coordinates $\textbf{x}$ is $\mathcal{G}_{\theta}$(\textbf{C}) such that $\mathcal{G}_{\theta}(\textbf{C}) \approx p_s(\textbf{C})$. Note that, while $p_s$ represents the value of the solution, $p_s(\textbf{C})$ denotes the solution operator. This operator is a function of the parameterized NURBS control points (\textbf{C}), which, in turn, are influenced by the shape of the rigid scatterer boundary ($\Gamma_i$).

It follows that the resulting PGI-DeepONet can simultaneously enforce the physics of the problem (physics-informed) and integrate NURBS-based geometry parameterization to capture arbitrary scatterer shapes (geometry-informed).   

\section{Network development} 
\label{ssec: Methodology}

In this section, we focus on the setup and implementation of the PGI-DeepONet with particular attention to the network architecture, the integration of geometry parameterization using NURBS, and the training procedure.

\subsection{PGI-DeepONet architecture}

We choose to develop a deep residual Network or ResNet architecture \cite{he2016deep} for the sub neural networks of the PGI-DeepONet. While the ability of ResNet to address vanishing gradient issues for data-driven DNNs is empirically proven \cite{szegedy2017inception}, recent studies on physics-driven DNNs \cite{qin2019data, karumuri2020simulator} highlight the role of ResNet in improving training efficiency and prediction accuracy. 

\begin{figure}[h!]
	\centering
	\includegraphics[width=1.0\linewidth]{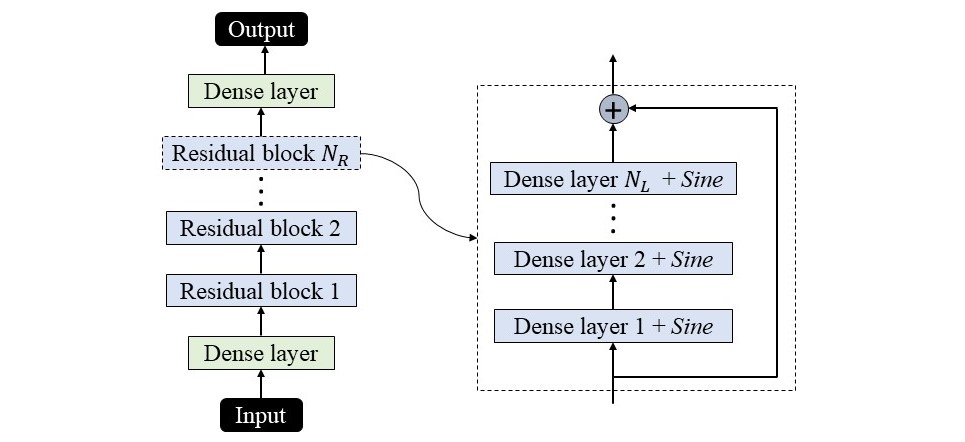}
	\caption{Schematic of the deep ResNet architecture with $N_R$ residual blocks, where each residual block contains $N_L$ linear layers with \textit{Sine} activation function and $n_w$ neurons (see dotted box). Note that the input and output of the ResNet architecture signifies the input and output of the branch and trunk networks.}
	\label{fig: ResNet}
\end{figure}

We use a ResNet architecture with $N_R=5$ residual blocks, each containing $N_L=3$ linear layers made of $n_w=100$ neurons using \textit{Sine} activation functions as shown in Fig.~\ref{fig: ResNet}. A residual block is an arrangement of linear neural network layers such that the input to the residual block is combined with its output deeper within the block for better information flow to the next stage of the DNN (see Fig.~\ref{fig: ResNet}). 

Our ResNet architecture passes the primary input through a linear layer of 100 neurons, then, the transformed input is passed through the residual blocks to generate a vector of size $200 \times 1$ as the output. Although the overall ResNet architecture for both branch and trunk networks are the same, due to the variation in the input sizes, we create two sub ResNets called the \textit{Branch ResNet} with an input size $16 \times 1$ and the \textit{Trunk ResNet} with input size $2 \times 1$. Table~\ref{table:Net_Arch} summarizes the parameters used to develop the Branch ResNet and the Trunk ResNet. Thereafter, as per Eq.~\ref{eqn: G_operator}, a dot product between the outputs of Branch ResNet ($\beta$) and the Trunk ResNet ($\tau$) evaluates $\mathcal{G}_{\theta}$. Note that, although $p_s$ is a complex number with real and imaginary components, DNNs cannot directly deal with complex numbers; hence, $\mathcal{G}_{\theta}$ is evaluated to contain real and imaginary components separately. It follows that, each of the output vectors $\beta$ and $\tau$ are split into two vectors of size $100 \times 1$ such that the dot products result in two scalar values. These scalar values are further concatenated to create the vector $\mathcal{G}_{\theta}$ of size $2 \times 1$, such that the first element corresponds to the real part and the second element corresponds to the imaginary part of the solution.
Note that since our solution approximation $\mathcal{G}_{\theta}$ contains both real and imaginary parts, the MSE losses in Eq.~\ref{eqn: loss_PIDeepONet_details} are evaluated separately for the real and imaginary components. Furthermore, we leverage automatic differentiation \cite{baydin2018automatic} to calculate the gradients of $\mathcal{G}_{\theta}$ with respect to the Trunk ResNet input $\textbf{x}$ for the loss function.

\begin{table}[h!]
\setlength\tabcolsep{5pt}
	\begin{center}
		\begin{tabular}{ |c|c|c| } 
			\hline
			\hline
			\multirow{1}*{\textbf{Parameter}} & \textbf{Branch ResNet} & \textbf{Trunk ResNet}\\
			\hline
			\hline
			Input and Output dimensions - [dim$_{in}$, dim$_{out}$] & [16, 200] & [2, 200] \\
			\hline   
			Hidden layer neurons & 100 & 100 \\
            \hline   
			Number of residual blocks & 5 & 5 \\
			\hline        
			Number of layers in each residual block & 3 & 3 \\
			\hline
			Activation function & Sine & Sine \\
            \hline
		\end{tabular}
		\caption{Summary of key network parameters used to develop the PGI-DeepONet architecture.}
		\label{table:Net_Arch}
	\end{center}
\end{table}

\subsection{Geometry preparation using NURBS}

\begin{figure}[h!]
	\centering
	\includegraphics[width=1.0\linewidth]{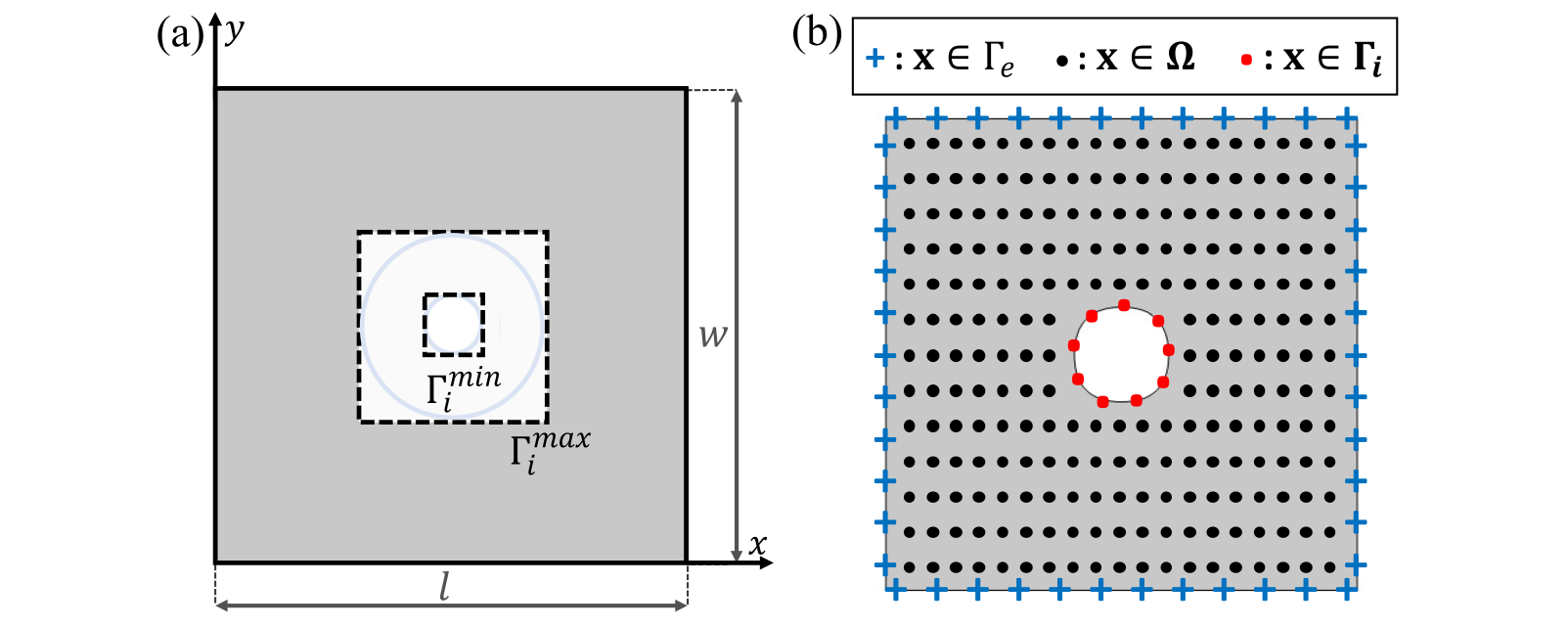}
	\caption{Schematic of (a) the 2D acoustic domain showing also the possible size range of internal boundaries ($\Gamma_i$). The size of the internal boundary can vary between a minimum and maximum value (dotted squares), i.e. $\Gamma_i \in [\Gamma^{min}_i, \Gamma^{max}_i]$. For example, $\Gamma_i$ could be a circle (in blue) with a radius between $\Gamma^{min}_i$ and $\Gamma^{max}_i$. In addition, the use of NURBS allows capturing many different arbitrary shapes contained within the size limits. (b) Coordinates used for training the physics-informed operator network on the external boundary ($\Gamma_e$), in the acoustic domain ($\Omega$), and on a sample circular internal boundary ($\Gamma_i$). }
	\label{fig: Acoustic_Domain_plots}
\end{figure}

To simulate rigid body acoustic scattering via PGI-DeepONet architecture, it is essential to supply the geometric features of the acoustic domain as input to the network. 
In this section, we discuss the input parameters for both the Branch and Trunk ResNet.
%
All the rigid scatterer shapes $\Gamma_i$ in this study are represented using nine NURBS control points $\textbf{C}=[\textbf{C}_0, \textbf{C}_1, ..., \textbf{C}_8]$ with a knot vector $t=[0,0,0,1/4,1/4,1/2,1/2,3/4,3/4,1,1,1]$, weights $\textbf{w}=[1,1,1,1,1,1,1,1,1]$, and quadratic B-spline basis functions. Table~\ref{table:NURBS} reports the key NURBS parameters. Moreover, as rigid scatterers are represented by closed curves, then $\textbf{C}_0 = \textbf{C}_8$. Therefore, any arbitrary shape can be accurately represented using a total of 16 NURBS parameters, consisting in 8 control points in the $x$-direction $[C^x_0, C^x_1, ...,C^x_7]$  and 8 control points in the $y$-direction $[C^y_0, C^y_1, ...,C^y_7]$. In addition, we limit the range of $\textbf{C}$ so to limit the possible design options. The range is carefully defined to ensure that, on one hand, the generated $\Gamma_i$ shapes form simple closed curves (no self-intersection or overlaps) within the acoustic domain, and on the other hand, they describe a wide design space. To this end, we define a lower and upper limit to the values of $\textbf{C}$ as reported in Table~\ref{table:NURBS}. Based on this range, the smallest possible shape is indicated by $\Gamma^{min}_i$ and the largest possible shape by $\Gamma^{max}_i$. A visual representation depicting the range of $\Gamma_i$ with circular samples is shown in Fig.~\ref{fig: Acoustic_Domain_plots}(a). 

Recall that the primary goal of this study is to develop an operator capable of predicting $p_s$ for any arbitrary rigid scatterer $\Gamma_i$ within the selected range. To achieve this result, the PGI-DeepONet must map the functional spaces of NURBS parameterized $\Gamma_i$ to the corresponding pressure field $p_s$ within an acoustic domain integrated with the scatterer $\Gamma_i$. In order to facilitate this functional space representation, a set of $m$ NURBS parameterized shapes denoted as $\{\Gamma^k_i\}^m_{k=1}$ is given as input to the Branch ResNet.
The collection of shapes $\{\Gamma^k_i\}^m_{k=1}$ is generated by applying choosing a uniform data distribution for $\textbf{C}$. Each shape $\Gamma_i$ is represented using its corresponding NURBS parameters $\textbf{C}$. More specifically, each shape in $\{\Gamma^k_i\}^m_{k=1}$  is structured in the form of a vector $\textbf{C}(\Gamma^k_i)$ of size $16 \times 1$, where $k=1,2,...m$ and used as input to the Branch ResNet, as shown in Fig.~\ref{fig: Acoustic_PIDeepONet}.

While the Branch ResNet is used to extract the features associated with the NURBS represented $\Gamma_i$, the Trunk ResNet is responsible for receiving the acoustic domain coordinates at which $p_s$ (and therefore $\mathcal{G}_{\theta}$) must be calculated. The coordinates $\textbf{x}$ are evaluated within a $[0,1] \times [0,1]$ square acoustic domain as shown in Fig.~\ref{fig: Acoustic_Domain_plots}(a). For each input to the Branch ResNet, derived from $\{\Gamma^k_i\}^m_{k=1}$ shapes, the corresponding $\textbf{x}=(x,y)$ of size $2 \times 1$ is provided as input to the Trunk ResNet. Note that $\textbf{x}$ could be a coordinate either on the external boundary of the acoustic domain or on the rigid scatterer boundary (i.e. $\textbf{x} \in \Gamma_e \cup \Omega \cup \Gamma_i$), as shown in Fig.~\ref{fig: Acoustic_Domain_plots}(b).

\begin{table}[h!]
\setlength\tabcolsep{5pt}
	\begin{center}
		\begin{tabular}{ |c|c|c| } 
			\hline
			\hline
			\multirow{1}*{\textbf{Parameter}} &  \textbf{Symbol} & \textbf{Numerical value}\\
			\hline
			\hline
			Degree of NURBS curve & $\tilde{k}$ & 2 \\
			\hline   
			Number of control points & $n+1$ & 9 \\
            \hline   
			Number of knots & $\tilde{m}+1$ & 12 \\
			\hline        
			Knot vector & $t$ & [0, 0, 0, $\frac{1}{4}$, $\frac{1}{4}$, $\frac{1}{2}$, $\frac{1}{2}$, $\frac{3}{4}$, $\frac{3}{4}$, 1, 1, 1] \\
			\hline
   		Weights & $w$ &
            [1,1,1,1,1,1,1,1,1] \\
			\hline
			Control points ($m$)& $\textbf{C}=[C^x, C^y]$ & $|C^x| \in [0.05,0.15],~  |C^y| \in [0.05,0.15]$\\
			\hline
   	    Domain length ($m$)& $l$ & 1\\
			\hline   
   		Domain width ($m$)& w & 1\\
			\hline
		\end{tabular}
		\caption{Summary of key geometry parameters used in the NURBS-based description of the rigid scatterers.}
		\label{table:NURBS}
	\end{center}
\end{table}

\subsection{Network training}
\label{ssec: network_training}

The geometric data introduced in the previous section were used as input to train the PGI-DeepONet. The network is trained on a set of $m=1000$ NURBS shapes such that $\{\Gamma^k_i\}^{1000}_{k=1}$ and the corresponding coordinates $\textbf{x}$ with $N_{\Gamma_e}=200$, $N_{\Omega}=10,000$, and $N_{\Gamma_i}=200$ as the input to the Branch and Trunk ResNets, respectively. Unlike the conventional DeepONet, the PGI-DeepONet does not train on a labeled input-output dataset. Instead, the PGI-DeepONet learns to predict accurate $\mathcal{G}_{\theta}$ corresponding to the input $\Gamma_i$ by minimizing the governing physics-based loss functions outlined in Eq.~\ref{eqn: loss_PIDeepONet}. Alternatively, the overall PGI-DeepONet architecture with two sub ResNets was trained jointly to predict $\mathcal{G}_{\theta}$ such that $\mathcal{G}_{\theta} \approx p_s$  by determining an optimal set of network parameters $\theta^*=[\theta_b^*, \theta_t^*]$ through loss minimization. In addition, a new set of 50 NURBS shapes $\{\Gamma^k_i\}^{50}_{k=1}$ was generated to test the prediction accuracy of the trained PGI-DeepONet. 

\begin{figure}[h!]
	\centering
	\includegraphics[width=1.0\linewidth]{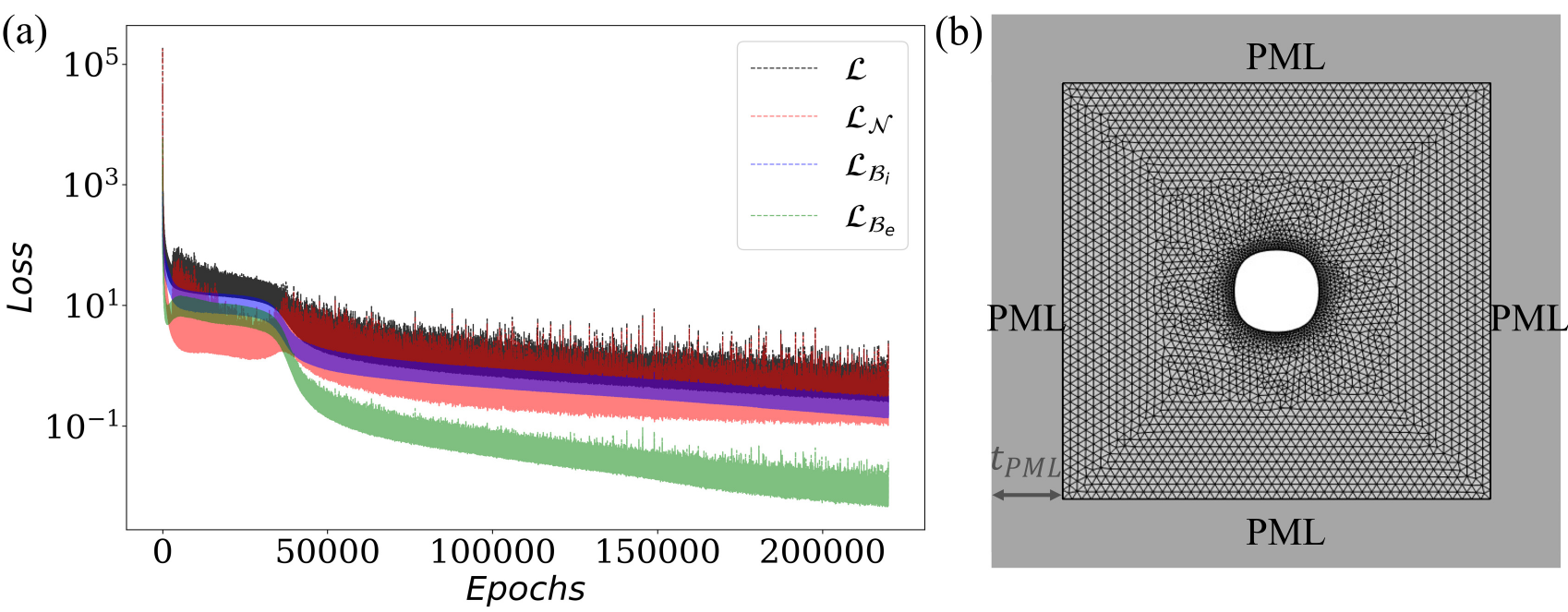}
	\caption{Schematic of (a) the convergence plot of the loss function $\mathcal{L}$ with training epochs. Note that this plot also records the variation in the individual loss components $\mathcal{L}_{\mathcal{N}}$, $\mathcal{L}_{\mathcal{B}_i}$, and $\mathcal{L}_{\mathcal{B}_e}$. (b) The meshed geometry for a sample circular $\Gamma_i$ with PML boundaries for finite element (FE) analysis.}
	\label{fig: Loss_plot}
\end{figure}

The PGI-DeepONet was trained for 220,000 epochs using the Adam optimizer with a learning rate of 5e-4 and with mini-batches of training samples in $\Gamma_i$, $\Omega$, and $\Gamma_e$. The network was trained until the loss function $\mathcal{L}$ converges as shown in Fig.~\ref{fig: Loss_plot}(a). The total training time was $30.6~hrs$. The PGI-DeepONet model was implemented in Python 3.8 using Pytorch API on a NVIDIA A100 Tensor Core GPU with 80GB memory.

\section{Numerical Results}
\label{ssec: Results}

In order to explore the performance of the PGI-DeepONets, two distinct scenarios were considered: 1) the same scatterer shape (i.e. circles) with varying sizes, and 2) arbitrary scatterer shapes with different sizes. In the following, the first scenario is referred to as NURBS-based circular shapes, and the second as NURBS-based arbitrary shapes. Finally, the computational time associated with the trained PGI-DeepONet simulations was also investigated.

To assess the quality of the results produced by the PGI-DeepONet, we compare the predicted pressure field $\mathcal{G}_{\theta}$ with a ground truth generated via finite element simulations. The acoustic scattering problem was simulated by COMSOL Multiphysics$\textsuperscript{\textregistered}$. The infinite acoustic domain was approximated by using absorbing boundary conditions applied all around the domain and implemented via a perfectly matched layer (PML) (Fig.~\ref{fig: Loss_plot}(b)). The following metrics are introduced to assess the overall performance of the PGI-DeepONet
\begin{enumerate}
    \item \textbf{Relative $L_2$ error:} The relative prediction quality is estimated, in an average sense, by
    \begin{equation}
        L_2 = \frac{||p_s - \mathcal{G}_{\theta}||_2}{|| p_s ||_2} 
    \end{equation}
    where, $p_s$ is the ground truth pressure field, $\mathcal{G}_{\theta}$ is the predicted pressure field, and $||.||_2$ is the standard Eucledian norm. 

    \item \textbf{$R^2$-score:} The coefficient of determination also called the $R^2$-score is used to assess the accuracy of the network prediction. The $R^2$-score measures the average variation in the predicted field ($\mathcal{G}_{\theta}$) with respect to the ground truth field ($p_s$) as follows
    \begin{equation}
        R^2 = 1- \frac{\sum^N_{i=1} (p^i_s - \mathcal{G}^i_{\theta})^2}{\sum^N_{i=1} (p^i_s - \bar{p}_s)^2} 
    \end{equation}
    where, $p^i_s$ and $\mathcal{G}^i_{\theta}$ are the ground truth pressure value and the predicted pressure value, respectively, at the $i^{th}$ location in the corresponding fields. Additionally, $N$ represents the number of locations where the pressure fields are compared. The maximum score of $R^2_{max}=1$ is achieved when the prediction matches exactly with the ground truth. Hence, the closer the $R^2$-score is to $R^2_{max}$, the higher the prediction accuracy.
    
    \item \textbf{Point-wise error:} The point-wise prediction quality of the pressure field is assessed by
    \begin{equation}
        \mathcal{E}_p = |p_s - \mathcal{G}_{\theta} |
    \end{equation}
    where $\mathcal{E}_p$ measures the relative error in percentage. This expression can also handle cases where the ground truth value is zero.
\end{enumerate}

\subsection{Performance assessment for NURBS-based circular shapes}
\label{ssec: NURBS_circular_shapes}

The performance of the trained PGI-DeepONet is evaluated using a test dataset different from the one used for training and containing 50 rigid scatterer shapes that were not part of the training dataset. Note that as the assessment specifically focuses on the prediction accuracy for circular rigid scatterers, the test dataset contains circular shapes of different radius.

\begin{figure}
	\centering
	\includegraphics[width=1.0\linewidth]{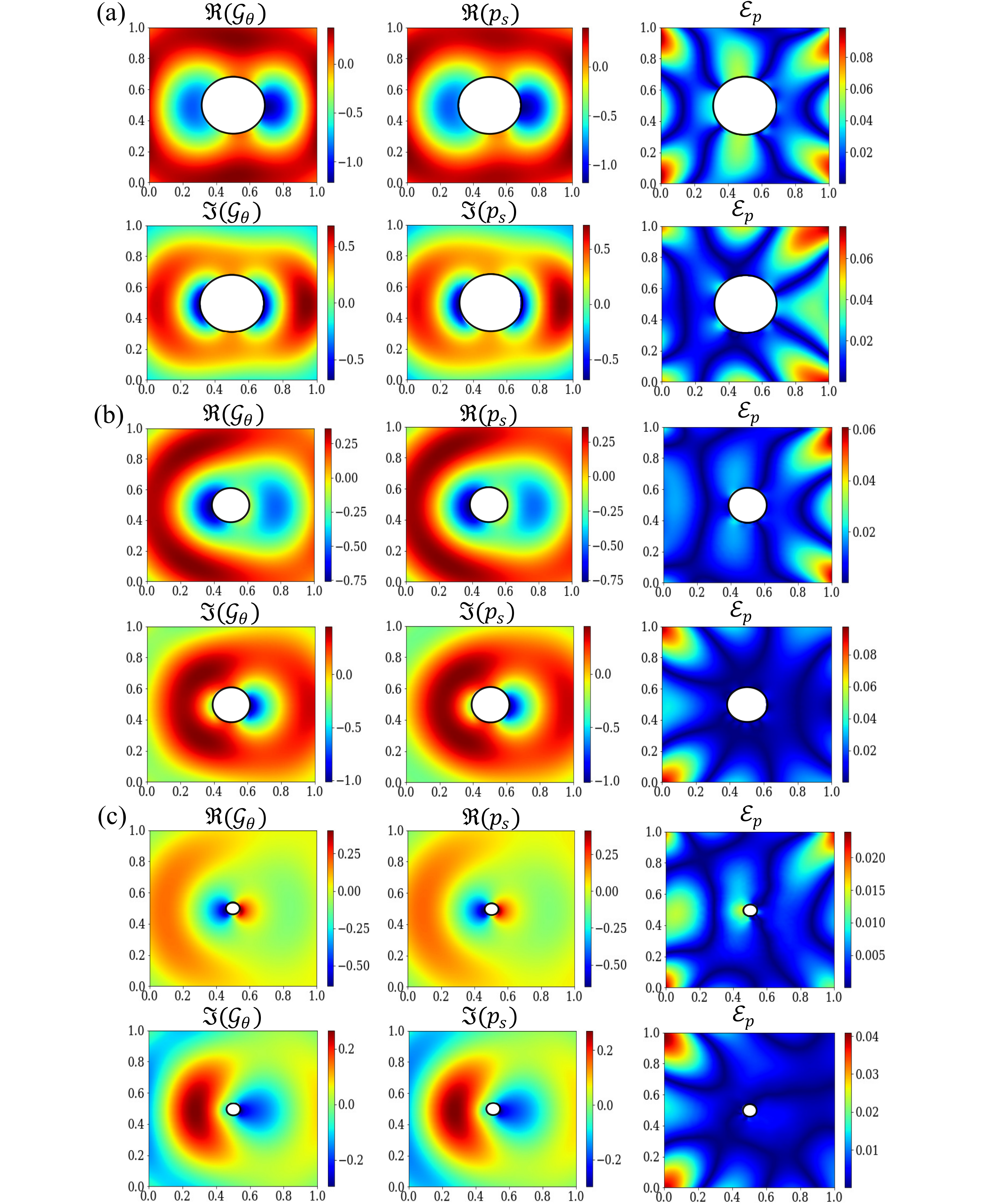}
	\caption{Field maps of the PGI-DeepONet predicted scattered acoustic pressure ($\mathcal{G}_{\theta}$), FE ground truth scattered acoustic pressure ($p_s$), and the point-wise error $\mathcal{E}_p$ between the prediction and ground truth. The real and imaginary pressure components are plotted for three different circles with radius: (a) $r=0.19m$, (b) $r=0.12m$, and (c) $r=0.05m$. }
	\label{fig: Results_circles}
\end{figure}
\begin{figure}
	\centering
	\includegraphics[width=1.0\linewidth]{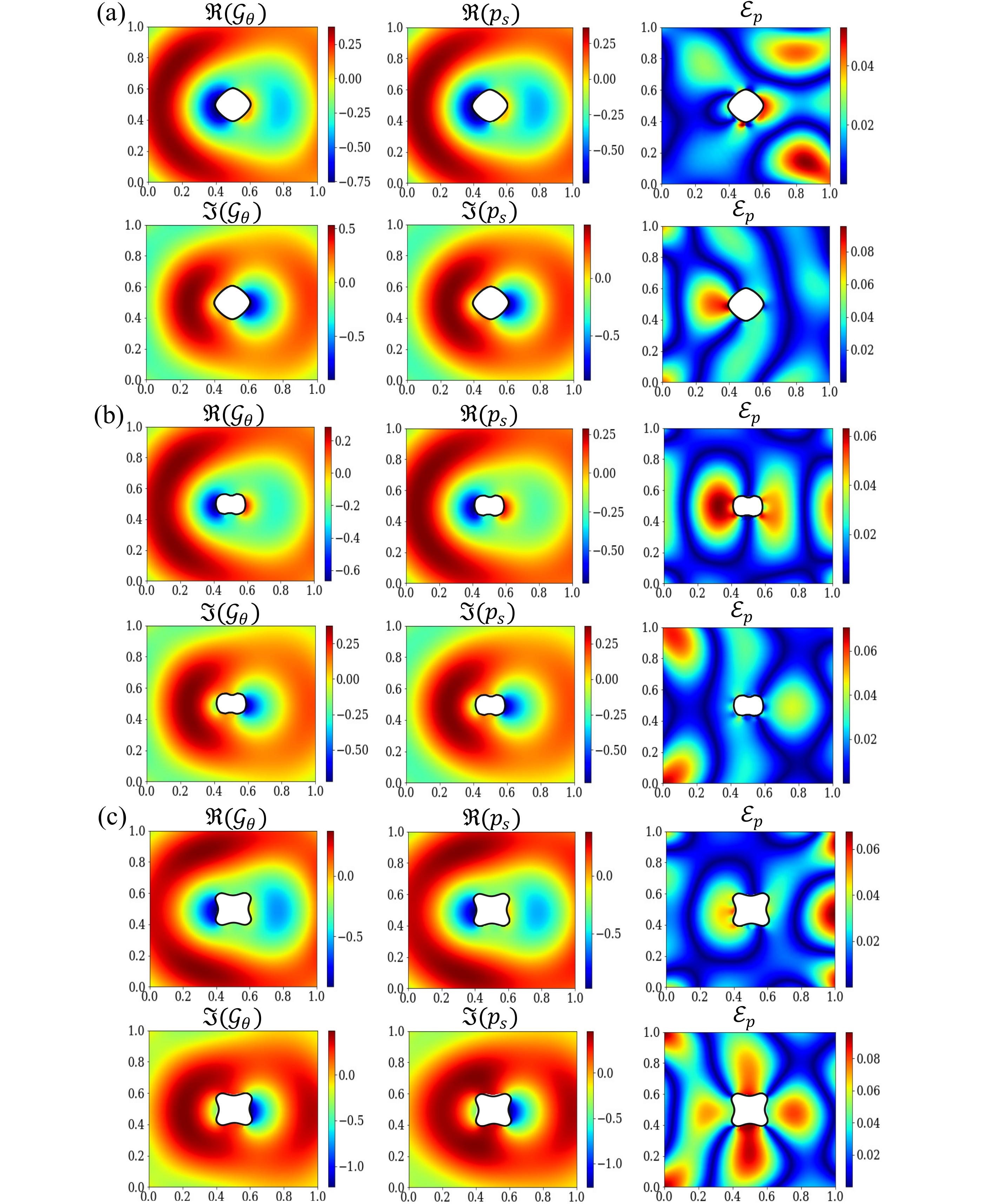}
	\caption{Field maps of the PGI-DeepONet predicted scattered acoustic pressure ($\mathcal{G}_{\theta}$), FE ground truth scattered acoustic pressure ($p_s$), and the point-wise error $\mathcal{E}_p$ between the prediction and ground truth. The real and imaginary pressure components are plotted for three arbitrary shapes: (a) shape-1 (skewed diamond shape), (b) shape-2 (peanut shape), and (c) shape-3 (star shape).}
	\label{fig: Results_arbitrary}
\end{figure}
Fig.~\ref{fig: Results_circles} compares the real and imaginary pressure fields predicted by the trained PGI-DeepONet for circular scatterers of different sizes. Specifically, the plot compares the predictions with the FE ground truth for three circular rigid scatterers of radius $r=0.19m$, $r=0.12m$, and $r=0.05m$. While the plots highlight the point-wise error $\mathcal{G}_{\theta}$, the calculated average was found to be $[L_2, R^2]_{r=0.19m}=[0.0852, 0.9904]$, $[L_2, R^2]_{r=0.12m}=[0.0724, 0.9937]$, and $[L_2, R^2]_{r=0.05m}=[0.0721, 0.9943]$. Moreover, the mean relative prediction error ($\bar{L}_2$) and mean prediction accuracy ($\bar{R}^2$) calculated across all the samples in the NURBS-based circular scatterers test dataset were found to be $[\bar{L}_2, \bar{R}^2] = [0.0781, 0.9926]$. These metrics illustrate the ability of PGI-DeepONet to accurately predict the scattered pressured field in an acoustic domain embedded with rigid circular scatterers of arbitrary radius. Note that while the present scenario was simulated at a frequency of $500Hz$, the PGI-DeepONet architecture can be retrained to accommodate different frequencies. The training process can remain relatively straightforward for lower frequencies, but utilization of optimized hyperparameters and training points might be necessary for higher frequencies.

\subsection{Performance assessment for NURBS-based arbitrary shapes}

Similar to the previous discussion, the performance of the trained PGI-DeepONet is evaluated using another test dataset consisting of 50 arbitrary rigid scatterer shapes. As this evaluation specifically focuses on the prediction accuracy for arbitrary rigid scatterers, the test dataset contains NURBS represented geometries of different shapes and sizes.

In Fig.~\ref{fig: Results_arbitrary}, we compare the real and imaginary pressure fields predicted by the trained PGI-DeepONet for scatterers of arbitrary shape and size. The figure highlights the point-wise error $\mathbf{e_{rel}}$ by comparing the PGI-DeepONet predictions with the FE ground truth for three arbitrary rigid scatterer shapes marked shape-1, shape-2, and shape-3. In addition, the average error was found to be $[L_2, R^2]_{shape-1}=[0.1055, 0.9873]$, $[L_2, R^2]_{shape-2}=[0.1273, 0.9827]$, and $[L_2, R^2]_{shape-3}=[0.1208, 0.9815]$. 
Moreover, the mean relative prediction error and mean prediction accuracy calculated across all the samples in the test dataset with NURBS-based arbitrary scatterers were found to be $[\bar{L}_2, \bar{R}^2] = [0.1468, 0.9713]$. This illustrates the ability of PGI-DeepONet to accurately predict the scattered pressured field in an acoustic domain embedded with arbitrary rigid scatterers. In addition, the results also indicate the ability of NURBS-based parameterization to capture a wide variety of shapes using only 16 geometry parameters. 

\begin{figure}[h!]
	\centering
	\includegraphics[width=1.0\linewidth]{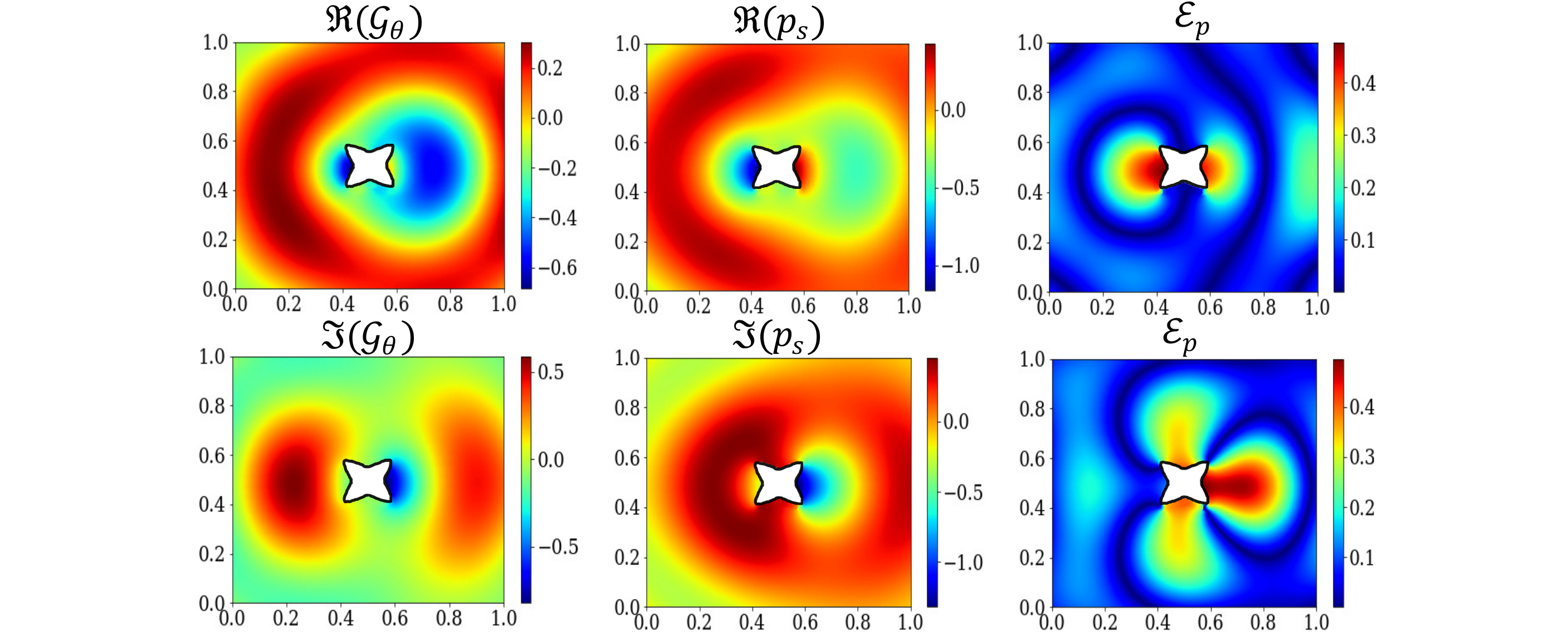}
	\caption{Field maps of the PGI-DeepONet predicted scattered acoustic pressure ($\mathcal{G}_{\theta}$), FE ground truth scattered acoustic pressure ($p_s$), and the the point-wise error $\mathcal{E}_p$ between the prediction and ground truth. The real and imaginary pressure components are plotted for an arbitrary shape with sharp edges.}
	\label{fig: Results_HighEr}
\end{figure}

Although the network is capable of predicting the scattered pressure field over a range of arbitrary shapes, the previous predictions for NURBS-based circular shapes (\S\ref{ssec: NURBS_circular_shapes}) had overall higher accuracy. This difference can be directly attributed to the ability of operators to easily learn function-to-function mapping for the same shape of different sizes. 
However, despite the small reduction in prediction accuracy, the integration of parsimonious shape representation using NURBS enables the PGI-DeepONet to approximate accurately the scattered fields across a wide variety of shapes.

Further, we also analyze the performance of PGI-DeepONet when predicting the pressure field for shapes with sharp edges, as shown in Fig.~\ref{fig: Results_HighEr}. This analysis specifically highlights significantly higher point-wise prediction errors $\mathcal{E}_p$ for a shape characterized by sharp edges. Our study indicates that the higher prediction errors for shapes with sharp edges can be directly attributed to the training approach of PGI-DeepONet. The PGI-DeepONet is trained to enforce the boundary equations on discretized training points on the rigid scatterer shapes. Since we aim to enable PGI-DeepONet to learn an operator for a broad range of shapes and sizes, we adopt a uniform shape discretization that can effectively represent the majority of shapes without a substantial increase in the computational cost for training. However, this uniform discretization lacks the precision to capture the details of shapes with sharp edges, hence resulting in fewer training points along these sharp edges and, finally, affecting the prediction accuracy as observed in the results in Fig.~\ref{fig: Results_HighEr}.

\subsection{Computational time}

While the previous sections studied the performance of the PGI-DeepONet in terms of its accuracy, another key attribute of the trained operator learning network is its ability to rapidly approximate the system's response. This section evaluates the computational time required by the trained PGI-DeepONet to approximate physically-consistent scattered acoustic fields for arbitrary scatterer shapes.
We compare the computational time employed by either the PGI-DeepONet or the FEM to calculate the same acoustic scattering problems. Note that all FEM simulations are obtained using the commercial software COMSOL Multiphysics$\textsuperscript{\textregistered}$, hence representing state-of-the-art in terms of FE based computations. 

The total computational time required by the PGI-DeepONet to perform the simulations can be broadly separated into the time needed to either train or predict. Notably, unlike other data-driven DNNs and DeepONets, PGI-DeepONet does not incur a data generation cost as it requires no labeled data for training. According to the details for network training discussed in \S\ref{ssec: network_training}, the training time of the PGI-DeepONet (for the entire range of selected arbitrary shapes) was approximately 30.6 $hrs$. 
Further, as introduced earlier, understanding the computational time required for network predictions is of particular importance. In the following, we evaluate and compare the computational time associated with individual network predictions as well as FEM calculations for equivalent scenarios.  

\begin{figure}[h!]
	\centering
	\includegraphics[width=1.0\linewidth]{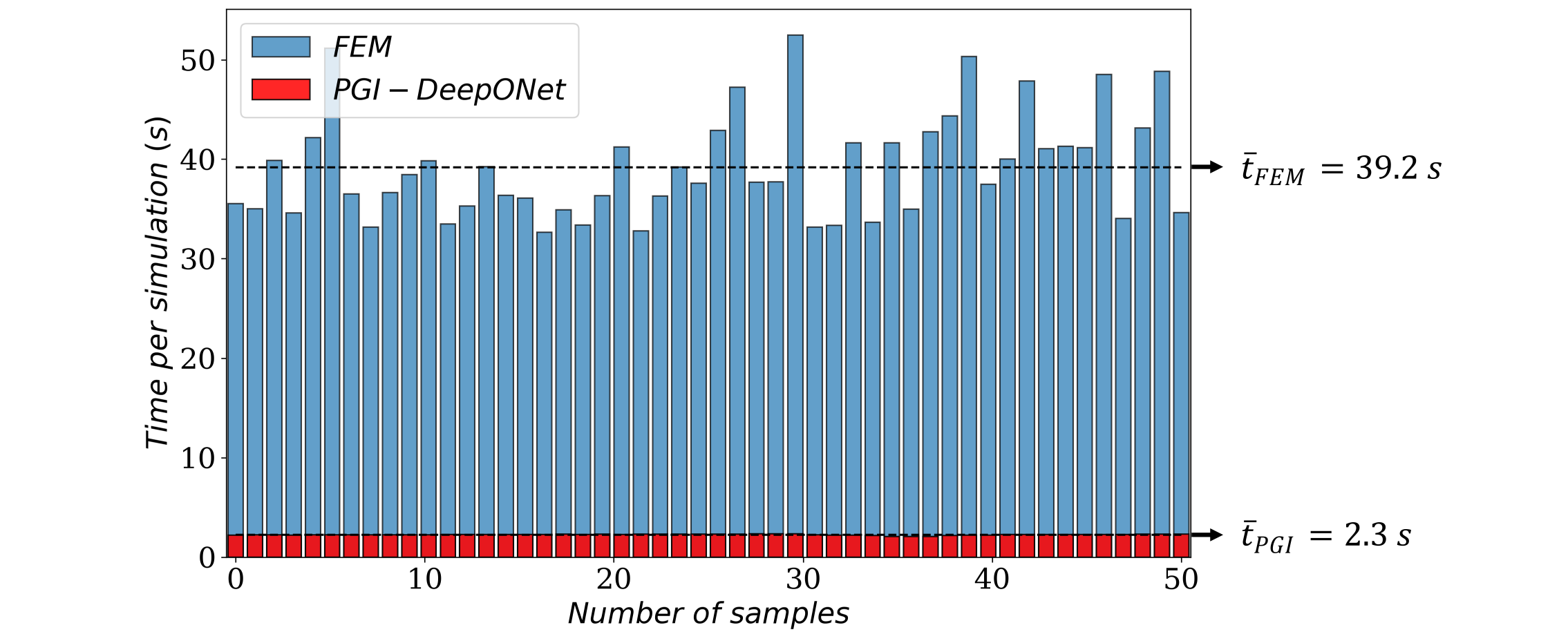}
	\caption{Bar chart comparing the computational time taken for each simulation. Results for 50 arbitrary-shaped scatterers are provided using FEM and trained PGI-DeepONet. The average computational time (dotted lines) for the FEM $\bar{t}_{FEM}$ and the PGI-DeepONet $\bar{t}_{PGI}$ over the 50 simulations are also reported.}
	\label{fig: Time_comparison}
\end{figure}

To this end, we compare the computational time per simulation $t$ for FEM and trained PGI-DeepONet across 50 simulations of varying scatterer shapes as shown in Fig.~\ref{fig: Time_comparison}. 
%
The study (see Fig.~\ref{fig: Time_comparison}) indicates that the average computational time for an arbitrary scatterer shape simulation using FEM is $\bar{t}_{FEM}=39.2~s$. On the other hand, the average computational time for an arbitrary scatterer shape simulation using trained PGI-DeepONet is $\bar{t}_{PGI}=2.3~s$. The direct comparison between the different $t$ values indicates the ability of the trained PGI-DeepONet to simulate scattered pressure fields for arbitrary scatterer geometries more than an order of magnitude faster ($\approx 17 \times$ faster) compared to the traditional forward solvers. Thus, the results highlight the ability of the trained PGI-DeepONet to serve as an efficient forward solver. It is also important to highlight that, compared to the prediction time $t$, the PGI-DeepONet training is computationally expensive. However, note that this is a one-time cost and the trained PGI-DeepONet can predict physically consistent solutions at an average computational time $\bar{t}_{PGI}$ for a continuous range of shapes (ideally an infinite number of shapes within the range) that can be captured by the integrated NURBS parameterization.

It is also key to observe that, while the computational time for traditional approaches (such as FEM) typically scales with the complexity of the problem (e.g. higher frequencies and complex geometries) due to its mesh dependence, the trained PGI-DeepONet can simulate these cases with marginal increase in prediction time due to its reduced dependence on domain discretization. For completeness, it should be mentioned that an increase in the complexity of the problem will likely increase the training time of the network. However, as mentioned above, the training is performed only once for a given range of shapes; it follows that, if the computational cost of the training process is seen relative to the number of predictions performed, this cost decreases progressively as the network is used to produce more predictions.

Lastly, both the FEM and trained PGI-DeepONet computations were evaluated using 10 processor cores of a high performance computational node equipped with $2\times$AMD EPYC 7763 "Milan" CPUs @ 2.2GHz with 256GB memory. 

\section{Conclusions}
\label{ssec: Conclusion}

This study introduced a geometry-aware and physics-driven neural operator network denominated physics- and geometry-informed deep operator network (PGI-DeepONet). The PGI-DeepONet was specifically applied and its performance assessed in the context of rigid body acoustic scattering problems. Existing approaches, including FE methods and PINNs, require evaluating the model every time the problem's conditions (e.g. boundary conditions, loads, and geometry) are changed. A trained PGI-DeepONet addresses this limitation by developing a solution operator that maps a continuous range of NURBS-represented scatterer shapes to their corresponding scattered pressure fields. The trained model improves the efficiency of the forward simulation required to approximate the scattered pressure field by orders of magnitude compared to finite element analysis. Although PGI-DeepONet was developed for rigid body scattering applications, the overall architecture is very general and can be adapted to different engineering problems involving computational domains of variable shape and size. The key attribute of this physics-informed neural operator network is its ability to capture geometries of arbitrary shapes and sizes using an integrated NURBS-based geometry parameterization. In addition, the physics-driven approach eliminates the need for labeled datasets during training. Instead, the model is trained to approximate the solution operator by enforcing both the governing equation and the boundary conditions of the acoustic scattering problem.

Further, we assessed the performance of the model by quantitatively comparing the accuracy of the PGI-DeepONet predictions against a ground truth obtained via FE simulations. The analysis highlighted the remarkable ability of the network to accurately approximate the scattered pressure field for scatterers having arbitrary shapes. In addition to emphasising the effectiveness of NURBS-based geometry parameterization to achieve versatile shape representation, these results also underscore the importance of physics-informed learning in order to generalize the approximation performance of the proposed network model.

\bigskip

\noindent \textbf{Competing interests}

The authors declare no competing interest.\\

\noindent \textbf{Acknowledgments}

This work was supported by the Laboratory Directed Research and Development program at Sandia National Laboratories, a multimission laboratory managed and operated by National Technology and Engineering Solutions of Sandia LLC, a wholly owned subsidiary of Honeywell International Inc. for the U.S. Department of Energy’s National Nuclear Security Administration under contract DE-NA0003525.

\bibliographystyle{unsrt} 
\bibliography{Finaldraft}

\begin{thebibliography}{10}

\bibitem{karniadakis2021physics}
GE~Karniadakis, IG~Kevrekidis, L~Lu, P~Perdikaris, S~Wang, and L~Yang.
\newblock Physics-informed machine learning.
\newblock {\em Nature Reviews Physics}, 3:422--440, 2021.

\bibitem{karpatne2022knowledge}
A~Karpatne, R~Kannan, and V~Kumar.
\newblock {\em Knowledge Guided Machine Learning: Accelerating Discovery Using
  Scientific Knowledge and Data}.
\newblock CRC Press, 2022.

\bibitem{white2019multiscale}
DA~White, WJ~Arrighi, J~Kudo, and SE~Watts.
\newblock Multiscale topology optimization using neural network surrogate
  models.
\newblock {\em Computer Methods in Applied Mechanics and Engineering},
  346:1118--1135, 2019.

\bibitem{zhang2021multi}
X~Zhang, F~Xie, T~Ji, Z~Zhu, and Y~Zheng.
\newblock Multi-fidelity deep neural network surrogate model for aerodynamic
  shape optimization.
\newblock {\em Computer Methods in Applied Mechanics and Engineering},
  373:113485, 2021.

\bibitem{pestourie2020active}
R~Pestourie, Y~Mroueh, TV~Nguyen, P~Das, and SG~Johnson.
\newblock Active learning of deep surrogates for {PDEs}: application to
  metasurface design.
\newblock {\em npj Computational Materials}, 6:1--7, 2020.

\bibitem{guo2021artificial}
K~Guo, Z~Yang, CH~Yu, and MJ~Buehler.
\newblock Artificial intelligence and machine learning in design of mechanical
  materials.
\newblock {\em Materials Horizons}, 8:1153--1172, 2021.

\bibitem{wu2022physics}
RT~Wu, M~Jokar, MR~Jahanshahi, and F~Semperlotti.
\newblock A physics-constrained deep learning based approach for acoustic
  inverse scattering problems.
\newblock {\em Mechanical Systems and Signal Processing}, 164:108190, 2022.

\bibitem{nair2023grids}
S~Nair, TF~Walsh, G~Pickrell, and F~Semperlotti.
\newblock {GRIDS-Net}: Inverse shape design and identification of scatterers
  via geometric regularization and physics-embedded deep learning.
\newblock {\em Computer Methods in Applied Mechanics and Engineering},
  414:116167, 2023.

\bibitem{liu2020multi}
Z~Liu, W~Cai, and ZQJ Xu.
\newblock Multi-scale deep neural network ({MscaleDNN}) for solving
  poisson-boltzmann equation in complex domains.
\newblock {\em Communications in Computational Physics}, 28(5), 2020.

\bibitem{mattheakis2019physical}
M~Mattheakis, P~Protopapas, D~Sondak, M~Di~Giovanni, and E~Kaxiras.
\newblock Physical symmetries embedded in neural networks.
\newblock {\em arXiv preprint arXiv:1904.08991}, 2019.

\bibitem{cai2020phase}
W~Cai, X~Li, and L~Liu.
\newblock A phase shift deep neural network for high frequency approximation
  and wave problems.
\newblock {\em SIAM Journal on Scientific Computing}, 42(5):A3285--A3312, 2020.

\bibitem{raissi2019physics}
M~Raissi, P~Perdikaris, and GE~Karniadakis.
\newblock Physics-informed neural networks: A deep learning framework for
  solving forward and inverse problems involving nonlinear partial differential
  equations.
\newblock {\em Journal of Computational physics}, 378:686--707, 2019.

\bibitem{shukla2020physics}
K~Shukla, PC~Di~Leoni, J~Blackshire, D~Sparkman, and GE~Karniadakis.
\newblock Physics-informed neural network for ultrasound nondestructive
  quantification of surface breaking cracks.
\newblock {\em Journal of Nondestructive Evaluation}, 39:1--20, 2020.

\bibitem{samaniego2020energy}
E~Samaniego, C~Anitescu, S~Goswami, VM~Nguyen-Thanh, H~Guo, K~Hamdia, X~Zhuang,
  and T~Rabczuk.
\newblock An energy approach to the solution of partial differential equations
  in computational mechanics via machine learning: {C}oncepts, implementation
  and applications.
\newblock {\em Computer Methods in Applied Mechanics and Engineering},
  362:112790, 2020.

\bibitem{raissi2020hidden}
M~Raissi, A~Yazdani, and GE~Karniadakis.
\newblock Hidden fluid mechanics: {L}earning velocity and pressure fields from
  flow visualizations.
\newblock {\em Science}, 367(6481):1026--1030, 2020.

\bibitem{jin2021nsfnets}
X~Jin, S~Cai, H~Li, and GE~Karniadakis.
\newblock {NSFnets} ({Navier-Stokes} flow nets): {P}hysics-informed neural
  networks for the incompressible {Navier-Stokes} equations.
\newblock {\em Journal of Computational Physics}, 426:109951, 2021.

\bibitem{zobeiry2021physics}
N~Zobeiry and KD~Humfeld.
\newblock A physics-informed machine learning approach for solving heat
  transfer equation in advanced manufacturing and engineering applications.
\newblock {\em Engineering Applications of Artificial Intelligence},
  101:104232, 2021.

\bibitem{yazdani2020systems}
A~Yazdani, L~Lu, M~Raissi, and GE~Karniadakis.
\newblock Systems biology informed deep learning for inferring parameters and
  hidden dynamics.
\newblock {\em PLoS computational biology}, 16(11):e1007575, 2020.

\bibitem{lu2021physics}
L~Lu, R~Pestourie, W~Yao, Z~Wang, F~Verdugo, and SG~Johnson.
\newblock Physics-informed neural networks with hard constraints for inverse
  design.
\newblock {\em SIAM Journal on Scientific Computing}, 43(6):B1105--B1132, 2021.

\bibitem{chen2020physics}
Y~Chen, L~Lu, GE~Karniadakis, and L~Dal~Negro.
\newblock Physics-informed neural networks for inverse problems in nano-optics
  and metamaterials.
\newblock {\em Optics express}, 28(8):11618--11633, 2020.

\bibitem{wang2021learning}
S~Wang, H~Wang, and P~Perdikaris.
\newblock Learning the solution operator of parametric partial differential
  equations with physics-informed {DeepONets}.
\newblock {\em Science advances}, 7(40):eabi8605, 2021.

\bibitem{lu2021learning}
L~Lu, P~Jin, G~Pang, Z~Zhang, and GE~Karniadakis.
\newblock Learning nonlinear operators via {DeepONet} based on the universal
  approximation theorem of operators.
\newblock {\em Nature machine intelligence}, 3(3):218--229, 2021.

\bibitem{di2021deeponet}
PC~Di~Leoni, L~Lu, C~Meneveau, G~Karniadakis, and TA~Zaki.
\newblock Deeponet prediction of linear instability waves in high-speed
  boundary layers.
\newblock {\em arXiv preprint arXiv:2105.08697}, 2021.

\bibitem{yin2022simulating}
M~Yin, E~Ban, BV~Rego, E~Zhang, C~Cavinato, JD~Humphrey, and GE~Karniadakis.
\newblock Simulating progressive intramural damage leading to aortic dissection
  using {DeepONet}: an operator--regression neural network.
\newblock {\em Journal of the Royal Society Interface}, 19(187):20210670, 2022.

\bibitem{zhu2023fourier}
M~Zhu, S~Feng, Y~Lin, and L~Lu.
\newblock Fourier-{DeepONet}: Fourier-enhanced deep operator networks for full
  waveform inversion with improved accuracy, generalizability, and robustness.
\newblock {\em arXiv preprint arXiv:2305.17289}, 2023.

\bibitem{moya2023deeponet}
C~Moya, S~Zhang, G~Lin, and M~Yue.
\newblock Deeponet-grid-uq: A trustworthy deep operator framework for
  predicting the power grid’s post-fault trajectories.
\newblock {\em Neurocomputing}, 535:166--182, 2023.

\bibitem{koric2023data}
S~Koric and DW~Abueidda.
\newblock Data-driven and physics-informed deep learning operators for solution
  of heat conduction equation with parametric heat source.
\newblock {\em International Journal of Heat and Mass Transfer}, 203:123809,
  2023.

\bibitem{goswami2022physics}
S~Goswami, M~Yin, Y~Yu, and GE~Karniadakis.
\newblock A physics-informed variational {DeepONet} for predicting crack path
  in quasi-brittle materials.
\newblock {\em Computer Methods in Applied Mechanics and Engineering},
  391:114587, 2022.

\bibitem{wang2023long}
S~Wang and P~Perdikaris.
\newblock Long-time integration of parametric evolution equations with
  physics-informed deeponets.
\newblock {\em Journal of Computational Physics}, 475:111855, 2023.

\bibitem{ma1998nurbs}
W~Ma and JP~Kruth.
\newblock Nurbs curve and surface fitting for reverse engineering.
\newblock {\em The International Journal of Advanced Manufacturing Technology},
  14:918--927, 1998.

\bibitem{saini2017nurbs}
D~Saini, S~Kumar, and TR~Gulati.
\newblock {NURBS}-based geometric inverse reconstruction of free-form shapes.
\newblock {\em Journal of King Saud University-Computer and Information
  Sciences}, 29:116--133, 2017.

\bibitem{hughes2005isogeometric}
TJR Hughes, JA~Cottrell, and Y~Bazilevs.
\newblock Isogeometric analysis: {CAD}, finite elements, {NURBS}, exact
  geometry and mesh refinement.
\newblock {\em Computer methods in applied mechanics and engineering},
  194:4135--4195, 2005.

\bibitem{balu2019deep}
A~Balu, S~Nallagonda, F~Xu, A~Krishnamurthy, M~Hsu, and S~Sarkar.
\newblock A deep learning framework for design and analysis of surgical
  bioprosthetic heart valves.
\newblock {\em Scientific reports}, 9:1--12, 2019.

\bibitem{zhang2021simulating}
W~Zhang, S~Motiwale, M~Hsu, and MS~Sacks.
\newblock Simulating the time evolving geometry, mechanical properties, and
  fibrous structure of bioprosthetic heart valve leaflets under cyclic loading.
\newblock {\em Journal of the Mechanical Behavior of Biomedical Materials},
  123:104745, 2021.

\bibitem{chen1995universal}
T~Chen and H~Chen.
\newblock Universal approximation to nonlinear operators by neural networks
  with arbitrary activation functions and its application to dynamical systems.
\newblock {\em IEEE transactions on neural networks}, 6(4):911--917, 1995.

\bibitem{piegl1996nurbs}
L~Piegl and W~Tiller.
\newblock {\em The NURBS book}.
\newblock Springer Science \& Business Media, 1996.

\bibitem{he2016deep}
K~He, X~Zhang, S~Ren, and J~Sun.
\newblock Deep residual learning for image recognition.
\newblock In {\em Proceedings of the IEEE conference on computer vision and
  pattern recognition}, pages 770--778, 2016.

\bibitem{szegedy2017inception}
C~Szegedy, S~Ioffe, V~Vanhoucke, and A~Alemi.
\newblock Inception-v4, inception-resnet and the impact of residual connections
  on learning.
\newblock In {\em Proceedings of the AAAI conference on artificial
  intelligence}, volume~31, 2017.

\bibitem{qin2019data}
T~Qin, K~Wu, and D~Xiu.
\newblock Data driven governing equations approximation using deep neural
  networks.
\newblock {\em Journal of Computational Physics}, 395:620--635, 2019.

\bibitem{karumuri2020simulator}
S~Karumuri, R~Tripathy, I~Bilionis, and J~Panchal.
\newblock Simulator-free solution of high-dimensional stochastic elliptic
  partial differential equations using deep neural networks.
\newblock {\em Journal of Computational Physics}, 404:109120, 2020.

\bibitem{baydin2018automatic}
AG~Baydin, BA~Pearlmutter, AA~Radul, and JM~Siskind.
\newblock Automatic differentiation in machine learning: a survey.
\newblock {\em Journal of Machine Learning Research}, 18:1--43, 2018.

\end{thebibliography}

\end{document}